# Cooperative Distributed Robust Control of Modular Mobile Robots with Bounded Curvature and Velocity

Xiaorui Zhu, *Student Member,* Youngshik Kim, *Student Member,* and Mark A. Minor, *Member, IEEE/ASME*

***Abstract*— A novel motion control system for Compliant Framed wheeled Modular Mobile Robots (CFMMR) is studied in this paper. This type of wheeled mobile robot uses rigid axles coupled by compliant frame modules to provide both full suspension and enhanced steering capability without additional hardware. The proposed control system is developed by combining a bounded curvature-based kinematic controller and a nonlinear damping dynamic controller. In particular, multiple forms of controller interaction are examined. A two-axle scout CFMMR configuration is used to evaluate the different control structures. Experimental results verify efficient motion control of posture regulation.**

## I. Introduction

A distributed robust control algorithm with bounded control inputs (curvature and velocity) is developed for a mobile robot and applied to a final target, a two-axle CFMMR (Compliant Framed wheeled Modular Mobile Robot), Fig. 1. The CFMMR has uniqueness in modularity and simple structure to provide axle suspension and highly controllable steering capability without any additional hardware. Despite these attributes, the CFMMR provides new challenges in motion control [1, 2], sensor instrumentation [3], and data fusion [3].

Control of compliance in robotic systems has been predominant amongst flexible manipulators where oscillations are a primary concern [4, 5]. Compliance control in the CFMMR differs in very substantial ways. The compliant frames encounter large deflections (i.e., post-buckling) and these nonlinear compliance effects are considerable. Therefore, traditional motion controllers are not well suited for a CFMMR. Compliance amongst mobile robots cooperatively manipulating an object has also received attention [6, 7], but these efforts have focused only on motion planning and coordination issues and ignored dynamic motion control subject to nonholonomic constraints.

In recent years, much attention has been paid to the motion control of mobile robots. Some research focuses only on the purely *kinematic motion controllers* where the control input is velocity [1, 8]. Practically, however, the robot dynamics should be taken into account such that the controller can produce the desired velocity using wheel torque provided by the motors. Thus, some research has been oriented toward *dynamic motion control* using torque inputs applied to control of dynamic and kinematic models subject to nonholonomic constraints in order to improve tracking performance [9, 10]. These efforts have focused only on rigid mobile robots not interacting cooperatively with other robots.

For motion control of the CFMMR, we should consider compliant coupling (e.g. cooperation and bounded path curvature) between axle modules. Thus, a curvature-based bounded *kinematic motion controller* should be realized to specify individual axle motion such that the CFMMR executes the desired net motion. These individual axle motions then provide real-time reference inputs to a *dynamic motion controller*. Therefore, the CFMMR control system consists of two parts; (1) time invariant bounded *kinematic motion control* and (2) robust *dynamic motion control*.

In our previous work [2], a perfect model of the CFMMR was assumed in the dynamic controller. However, in reality, disturbances may not be negligible. Due to the unpredictable bounds of disturbances of the CFMMR dynamic model, a robust nonlinear damping control technique [11] is extended to our dynamic controller. The main contributions of this paper involve unifying bounded kinematic control and robust dynamic control. Each of these controllers are developed independently to resolve their own issues and then combined together in order to analyze their interaction. The structure of the paper follows. Modular models of the CFMMR are derived in Section 2. A *kinematic motion controller* is proposed in Section 3. A *dynamic motion controller* is proposed in Section 4. The two proposed controllers are combined in Section 5. The control algorithm is applied to a two-axle CFMMR to evaluate the

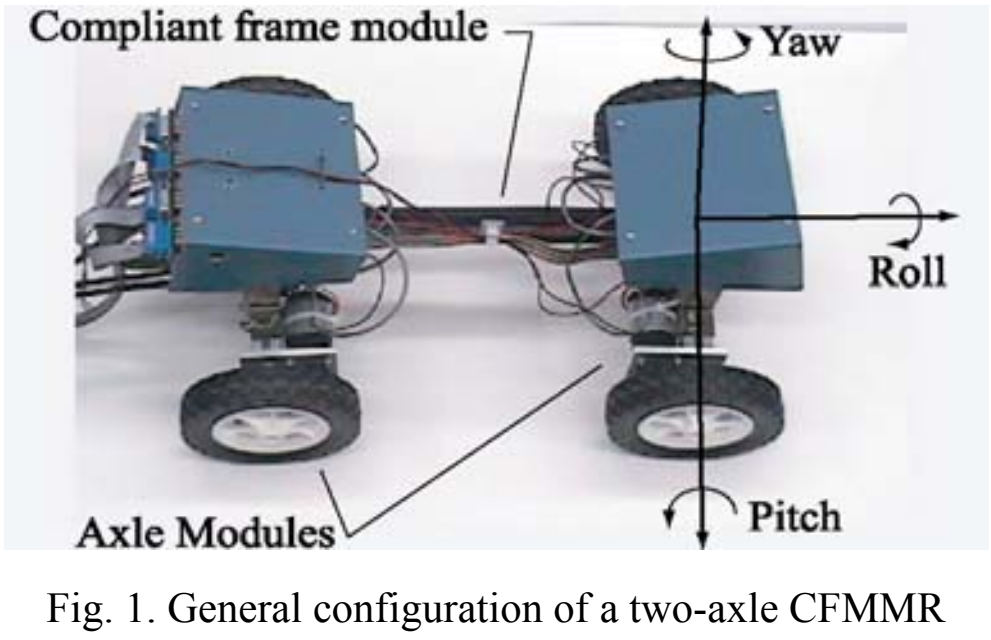

Fig. 1. General configuration of a two-axle CFMMR

The authors gratefully acknowledge support of this research by NSF Grant No. IIS-0308056.

X. Zhu is with the Dept. of Mechanical Engineering, University of Utah, Salt Lake City, UT 84112 USA (e-mail: xiaorui.zhu@utah.edu).

Y. Kim is with the Dept. of Mechanical Engineering, University of Utah, Salt Lake City, UT 84112 USA (e-mail: youngshik.kim@utah.edu).

M. A. Minor is with the Dept. of Mechanical Engineering, University of Utah, Salt Lake City, UT 84112 USA (phone: 801-587-7771; fax: 801-585-9826; e-mail: minor@mech.utah.edu).

performance in Section 6. Concluding remarks and future work are described in Section 7.

## II. Kinematic and Dynamic Models

### A. Generic Modeling Structure

Consider the $i^{th}$ axle of an $n$-axle CFMMR, let us define a fixed global reference frame $G(X_G, Y_G)$ and moving frames $f_i(x_i, y_i)$ attached to the points $C_i$ at the midpoint of the $i^{th}$ axle, where $i = 1,...,n$. At any instant, the $i^{th}$ axle module is rotating about the IC (Instantaneous Center), Fig. 2, such that the IC's projections onto the $x_i$ axes define point $C_i$. A module configuration vector $q_i = [X_i \quad Y_i \quad \phi_i]$ is then attached to this point for each axle. In order to describe this configuration within the context of the entire system, we assemble each of these module configuration vectors into a system configuration vector $Q = [q_1,...,q_n]^T$ where $Q \in R^{3n\times 1}$. It is then possible to assemble a system description of the form,

$$\mathbf{M}(Q)\ddot{Q} + \mathbf{V}(Q,\dot{Q})\dot{Q} + \mathbf{F}(\dot{Q}) + \mathbf{G}(Q) + \tau_d + \mathbf{F}_K(Q) = \mathbf{E}(Q)\tau - \mathbf{A}^T(Q)\lambda \tag{1}$$

where $\mathbf{M}(Q) \in R^{3n\times 3n}$ is a symmetric, positive definite inertia matrix assembled from the individual axle module inertia matrices. Assembling individual axle module dynamic characteristics into the system model, $\mathbf{V}(Q,\dot{Q}) \in R^{3n\times 3n}$ is the centripetal and Coriolis forces, $\mathbf{F}(\dot{Q}) \in R^{3n\times 1}$ denotes the friction, $\mathbf{G}(Q) \in R^{3n\times 1}$ is the gravitational vector, $\tau_d$ denotes bounded unknown disturbances including unstructured un-modeled dynamics, $\mathbf{E}(Q) \in R^{3n\times 2n}$ is the input transformation matrix, $\tau \in R^{2n\times 1}$ is the input torques, and $\lambda \in R^{n\times 1}$ is the vector of constraint forces. $\mathbf{A}(Q) \in R^{n\times 3n}$ is the global matrix associated with the nonholonomic constraints. Compliant frame forces are described by globally defined stiffness equations that are assembled into $\mathbf{F}_K(Q) \in R^{3n\times 1}$.

Considering nonholonomic constraints, let $\mathbf{S}(Q)$ be a full rank matrix formed by a set of smooth and linearly independent vector fields spanning the null space of $\mathbf{A}(Q)$ such that,

$$\mathbf{A}(Q)\mathbf{S}(Q) = \mathbf{0}. \tag{2}$$

Therefore we can find an auxiliary velocity vector $\mathbf{v}(t)$ such that for all time, $t$,

$$\dot{Q} = \mathbf{S}(Q)\mathbf{v}(t). \tag{3}$$

### B. Modular Dynamic Models

Considering the $i^{th}$ axle module, we let $\mathbf{V}_i(q_i, \dot{q}_i) = \mathbf{0}$ since centripetal and Coriolis forces of each axle are relatively small due to velocity magnitudes. Also, $\mathbf{G}_i(q_i) = \mathbf{0}$ assuming the robot is on a horizontal plane. The axle module matrices are:

$$\mathbf{M}_i(q_i) = \begin{bmatrix} m_i & 0 & 0 \\ 0 & m_i & 0 \\ 0 & 0 & J_i \end{bmatrix}, \mathbf{E}_i(q_i) = \frac{1}{r_w}\begin{bmatrix} \cos\phi_i & \cos\phi_i \\ \sin\phi_i & \sin\phi_i \\ -d & d \end{bmatrix}, \tag{4}$$

$$\text{and } \mathbf{S}_i(q_i) = \begin{bmatrix} \cos\phi_i & 0 \\ \sin\phi_i & 0 \\ 0 & 1 \end{bmatrix}. \tag{5}$$

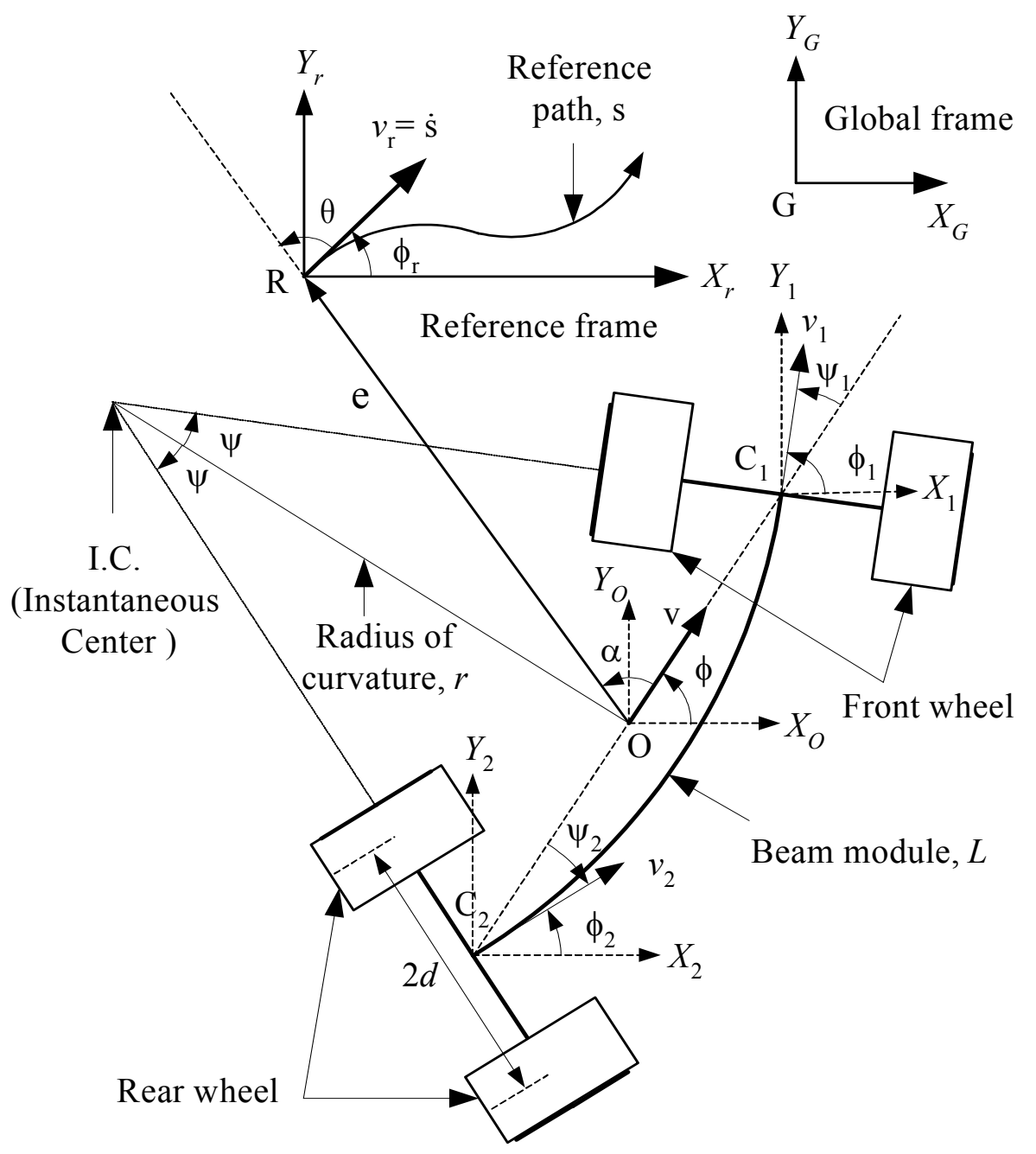

Fig. 2. General configuration of a two-axle CFMMR

The compliant frame force vector, $\mathbf{F}_K$, was developed based on the Finite Element Method (FEM) and the post-buckled frame element [2]. Assuming a serial configuration, (4) and (5) are assembled into (1) to form block diagonal system matrices.

## III. Curvature-based kinematic motion control

### A. Kinematic Model

The CFMMR has much more complex steering kinematics than unicycle type robots since it possesses independently steered axles that are compliantly coupled. However, our previous research [1, 12] shows that a two-axle CFMMR can be efficiently steered via a reduced equivalent kinematic model, which is derived from front and rear axle relative steering angles $\psi_1 = -\psi_2$, Fig. 2. Thus, the net position and orientation of the robot is described by an equivalent posture attached to point $O$ located at the center of $\overline{C_1C_2}$. For simplicity we use only a forward velocity.

Using the polar representation relative to posture $O$, the equivalent unicycle kinematics can be written in error coordinates,

$$\begin{aligned}\dot{e} &= -v\cos\alpha + v_r\cos\theta \\ \dot{\theta} &= v\frac{\sin\alpha}{e} - v_r\frac{\sin\theta}{e} - \dot{\phi}_r \\ \dot{\alpha} &= v\frac{\sin\alpha}{e} - v_r\frac{\sin\theta}{e} - \dot{\phi}\end{aligned} \tag{6}$$

where the variable $v$ represents the velocity of the coordinate frame $O$ moving in a heading $\phi$ relative to the global frame $G$. The subscript $r$ denotes the reference frame. That is, $v_r$ and $\phi_r$ are the reference velocity and the reference heading angle of the coordinate frame $R$, respectively.

### B. Curvature based geometric approach

In this research, the robot must proceed only on paths of bounded curvature imposed by kinematic configurations and traction forces. To command bounded curvature inputs for motion control, a curvature based geometric approach was presented [13]. In this approach, a smooth *path manifold* drives a robot to a target satisfying path curvature bounds with minimal control effort.

To obtain an explicit closed form expression, a circular path is employed as a predefined *path manifold*, Fig. 3. The robot velocity v converges to the reference velocity $v_r$ and the position error $e$ becomes zero as $\theta$ goes to zero. That is, posture regulation and path following of a robot can be achieved based on this geometric approach if one of the error variables converges to zero. To realize this circular path and constrain the motion of the system to the path, the *path manifold* is established by,

$$\begin{aligned}e &= r\sqrt{2(1-\cos 2\theta)} \\ \alpha &= -\theta\end{aligned} \tag{7}$$

where $r$ is the radius of a circular *path manifold*. Since $r$ is the inverse of curvature, $r$ should satisfy $r \geq 1/\kappa_{\max}$ where $\kappa_{\max}$ is the maximum allowed path curvature of the robot.

To analyze the *path manifold* in detail, differentiate (7):

$$\begin{aligned}\dot{e} &= 2r\dot{\theta}\cos(\theta)U(\theta) \\ \dot{\alpha} &= -\dot{\theta}\end{aligned}; \quad U(\theta) = \begin{cases}1, & \text{if } 0 \leq \theta \leq \pi \\ -1, & \text{if } -\pi \leq \theta < 0\end{cases}. \tag{8}$$

Referring to (6), (7) and (8), velocity expressions are,

$$\begin{aligned}v &= v_r - 2r\dot{\theta}U(\theta) \\ \dot{\phi} &= 2\dot{\theta} + \dot{\phi}_r \\ \dot{\phi}_r &= -\frac{v_r}{r}U(\theta)\ .\end{aligned} \tag{9}$$

These results verify that the origin is in equilibrium (i.e., $\dot{e} = \dot{\theta} = \dot{\alpha} = 0$) via the circular *path manifold*.

To achieve adequate maneuverability within the path curvature limit via the geometric approach, the robot must be placed initially outside of the convergence circle. In case of non-conforming initial conditions, we can resolve this by commanding the robot to move outside of the convergence circles and then maneuver using the new initial condition.

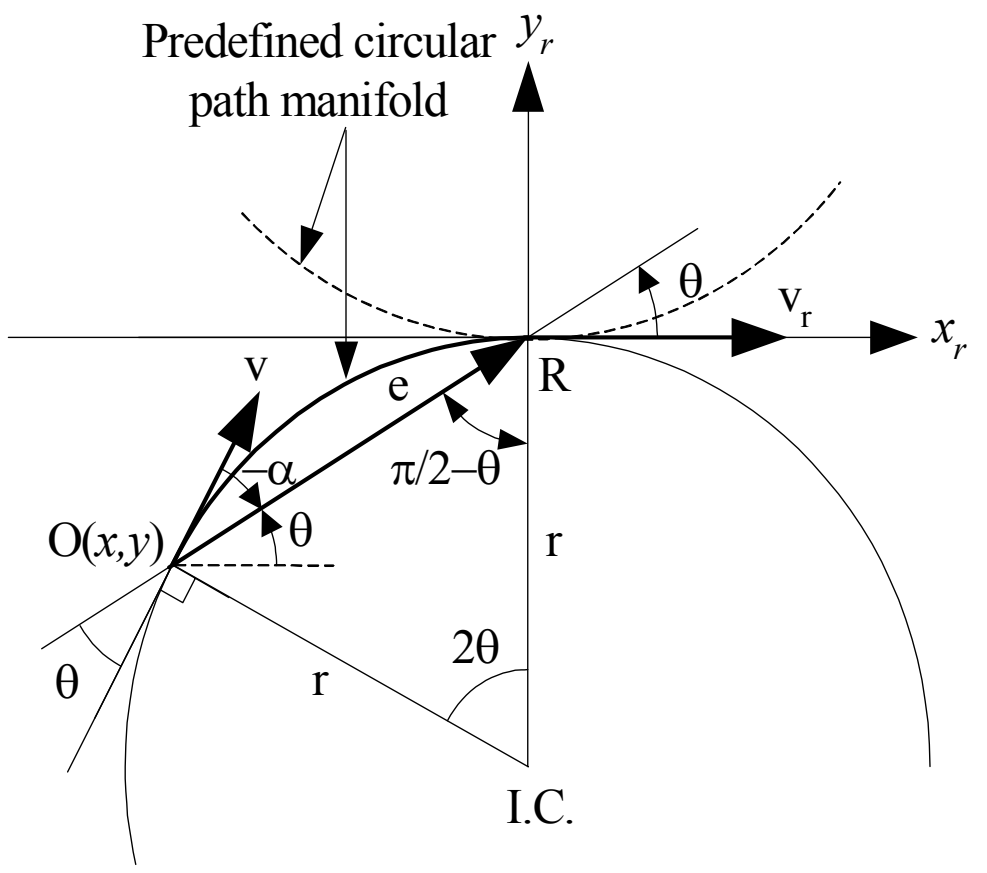

Fig. 3. Curvature based approach using a predefined circular *path manifold.*

### C. Smooth Time Invariant Bounded Control Law

A curvature based time invariant control law was developed to solve both posture regulation and path following problems simultaneously with bounded control inputs [13]. Using Lyapunov techniques the robot system will reach a *path manifold* in finite time and be driven to the equilibrium point along the *path manifold* asymptotically, similar to sliding mode control techniques.

To circumvent singularity at the origin, we introduce a sufficiently small perturbation, ε. Denoting $\zeta=1+\varepsilon$ and letting $z = e - r\sqrt{2}\sqrt{\zeta - \cos 2\theta} \geq 0$, quadratic Lyapunov candidate functions, $V_1$ and $V_2$ are determined as,

$$\begin{aligned}V_1 &\equiv \frac{1}{2}z^2 \\ V_2 &\equiv \frac{1}{2}(\theta+\alpha)^2\ .\end{aligned} \tag{10}$$

For bounded controllers and asymptotic stability, $\dot{V}_1$ and $\dot{V}_2$ should be a negative semi-definite function such that,

$$\begin{aligned}\dot{V}_1 &= z\dot{z} = -k_1 z\tanh(z) - O(\varepsilon) \leq 0 \\ \dot{V}_2 &= (\theta+\alpha)(2\dot{\theta} + \dot{\phi}_r - \dot{\phi}) = -k_2(\theta+\alpha)\tanh(\theta+\alpha) \leq 0\end{aligned} \tag{11}$$

where $O(\varepsilon)$ presents a disturbed positive definite function in the order of $\varepsilon$. Since $z \geq 0$ and $0 \leq v \leq v_{\max}$, the following inequalities must be true for a singularity-free controller, $v$,

$$-z(1-\cos\alpha)v_{\max} \leq \dot{V}_1 + k_1 z\tanh(z) - z(1-\cos\alpha)v \leq 0\ . \tag{12}$$

Solving for $v$ and $\dot{\phi}$ from (11) and (12), and using a curvature expression, $\kappa=\dot{\phi}/v$, the proposed control law is finally determined as,

$$v=\frac{\left\{\begin{array}{l}k_1 e\sqrt{\zeta-\cos 2\theta}\tanh(e-r\sqrt{2}\sqrt{\zeta-\cos 2\theta})\\+v_r e\cos\theta\sqrt{\zeta-\cos 2\theta}+v_r r\sqrt{2}\sin 2\theta(\sin\theta+\kappa_r e)\end{array}\right\}}{e\sqrt{\zeta-\cos 2\theta}+r\sqrt{2}\sin 2\theta\sin\alpha} \quad (13)$$

$$\kappa=\frac{\dot\phi}{v}=\frac{k_2\tanh(\theta+\alpha)+\frac{2}{e}(v\sin\alpha-v_r\sin\theta)-v_r\kappa_r}{v}.$$

This control law can solve both the posture regulation and the path following problems with bounded curvature and velocity. Referring to (6) and (13), $(e,\theta,\alpha)$ $\to$ $(r\sqrt{2\varepsilon},-\kappa_r r\sqrt{2\varepsilon}/2,\kappa_r r\sqrt{2\varepsilon}/2)\simeq(0,0,0)$, $(\dot e,\dot\theta,\dot\alpha)\to(0,0,0)$ $v\to v_r$ and $\dot\phi\to\dot\phi_r$ (i.e. $\kappa\to\kappa_r$) as $t\to\infty$ for sufficiently small perturbation, ε. Hence, the origin may be considered as the equilibrium point of the system (6) without loss of generality.

Further, given bounded control inputs $v_{\max}$ = 0.5 m/sec and $\kappa_{\max}$ = 3 m$^{-1}$ for the current two-axle CFMMR, optimization techniques and worst-case analysis were conducted to determine gains and smooth reference:

$$\begin{aligned}&k_1=0.2+0.3\left(1-\tanh\left(1.3/e\right)\right)\\&k_2=0.3/e(0)\tanh\left(1/2|\alpha|\right)+0.3\tanh\left(1/e\right)\\&v_r=v_{des}\tanh\left(0.1/e\right)\\&\kappa_r=\kappa_{des}\end{aligned} \quad (14)$$

where $v_{des}$ and $\kappa_{des}$ represent desired velocity and curvature of particular path segments, respectively.

## IV. Robust Dynamic Motion Control Design

The curvature-based *kinematic motion control* algorithms we discussed in Sect. III provide bounded time-varying reference trajectories. The nonlinear damping *dynamic controller* forces the robot to follow these reference trajectories [14].

### A. Structural Transformation of Single Axle Module

Considering the $i^{th}$ axle module, we rewrite the $i^{th}$ dynamic equations such that the regular backstepping form can be obtained:

$$\dot q_i=\mathbf{S}(q_i)\mathbf{v}_i \quad (15)$$

$$\bar{\mathbf{M}}_i\dot{\mathbf{v}}_i+\bar B_i\mathbf{v}_i+\bar\tau_{d,i}=\bar\tau_i \quad (16)$$

where $\bar{\mathbf{M}}_i=\mathbf{S}_i^T\mathbf{M}_i\mathbf{S}_i$, $\bar B_i=\mathbf{S}_i^T(\mathbf{M}_i\dot{\mathbf{S}}_i+B_i\mathbf{S}_i)$, $\bar\tau_i=\mathbf{S}_i^T\mathbf{E}_i\tau_i$, and $\bar\tau_{d,i}=\mathbf{S}_i^T(\tau_{d,i}+\mathbf{F}_{K,i}(q_i,q_j))$, $\mathbf{v}_i=\left[v_i\quad\omega_i\right]^T$, and $v_i$ and $\omega_i$ represent the linear and angular velocities of the point $C_i$. Here we assume $\mathbf{F}_i=B_i\dot q_i$, where $B_i$ consists of constant friction coefficients. The nonlinear component of the friction forces is included in the unknown disturbances, $\tau_{d,i}$.

The control objective is to derive a suitable $\tau_i(t)$ such that the CFMMR will track a smooth steering velocity,

$$\mathbf{v}_{c,i}(t)=\left[v_{c,i}\quad\omega_{c,i}\right]. \quad (17)$$

where the steering velocities serve as inputs for (15). The controller then provides trajectory tracking of the reference trajectories, $q_{r,i}$, with torque commands.

The reference trajectories $q_{r,i}$ can be solved from:

$$\dot q_{r,i}=S(q_{r,i})\mathbf{v}_{r,i}. \quad (18)$$

The error state model for tracking is then defined as,

$$e_i=\mathbf{R}_{\phi,i}\left(q_{r,i}-q_i\right) \quad (19)$$

where $q_{r,i}$ is the reference vector for the $i^{th}$ axle, $e_i\in R^{3\times1}$ is the error position vector for the $i^{th}$ axle and

$$e_i=\left[e_{X,i}\quad e_{Y,i}\quad e_{\phi,i}\right]^T. \quad (20)$$

As [15] shows, an alternative $\mathbf{v}_{c,i}$ is chosen as

$$\mathbf{v}_{c,i}=\begin{bmatrix}v_{r,i}\cos e_{\phi,i}+k_{X,i}e_{X,i}\\ \omega_{r,i}+k_{Y,i}v_{r,i}e_{Y,i}+k_{\phi,i}v_{r,i}\sin e_{\phi,i}\end{bmatrix}, \quad (21)$$

where $k_i=\left[k_{X,i}\quad k_{Y,i}\quad k_{\phi,i}\right]$ are positive constants for the $i^{th}$ axle. The velocity control law $\mathbf{v}_{c,i}$ is thus proven to make $e_i=0$ a stable equilibrium point using the Lyapunov function [15],

$$V_{1,i}(e_i)=\frac{1}{2}e^2{}_{X,i}+\frac{1}{2}e^2{}_{Y,i}+(1-\cos e_{\phi,i})/K_{y,i} \quad (22)$$

where $K_{y,i}$ is a positive constant and $V_{1,e}(e_i)$ is used in subsequent controller development.

### B. Properties and Assumptions of Single Axle Controller

There are several properties and assumptions that will be used in the following control design:

*Assumption: $\tau_{d,i}$ and $\mathbf{F}_{K,i}(q_i,q_j)$ are bounded.*

*Property1: $\bar{\mathbf{M}}_{\min,i}I_n\le\bar{\mathbf{M}}_i(q_i)\le\bar{\mathbf{M}}_{\max,i}I_n$, where $\bar{\mathbf{M}}_{\min,i}$, $\bar{\mathbf{M}}_{\max,i}$ are some positive scalar constants and $I_n$ is an $n\times n$ identity matrix.*

*Property 2: $\|\bar B_i(q_i)\|\le b_i$, where $b_i$ is a positive constant.*

*Property 3: $\bar{\mathbf{M}}_i$ is constant.*

*Property 4: $\dot{\mathbf{v}}_{c,i}=A_{1,i}\mathbf{v}_{c,i}+A_{2,i}\mathbf{v}_{r,i}+A_{3,i}\dot{\mathbf{v}}_{r,i}$, where $\|A_{1,i}\|$, $\|A_{2,i}\|$ and $\|A_{3,i}\|$ are bounded.*

### C. Nonlinear Damping Control Design of Single Axle Module

We will now extend the nonlinear damping control scheme to a single-axle CFMMR configuration [11]. Define the velocity error vector for each axle as,

$$\mathbf{e}_{c,i} = \begin{bmatrix} e_{v,i} \\ e_{\omega,i} \end{bmatrix} = \mathbf{v}_i - \mathbf{v}_{c,i} . \tag{23}$$

Differentiating (23) and substituting (16) yields,

$$\bar{\mathbf{M}}_i \dot{\mathbf{e}}_{c,i} = \bar{\tau}_i - \bar{B}_i \mathbf{v}_i - \bar{\tau}_{d,i} - \bar{\mathbf{M}}_i \dot{\mathbf{v}}_{c,i} . \tag{24}$$

Choose the Lyapunov candidate for the dynamic model (16) as

$$V_{2,i}(\mathbf{e}_{c,i}) = \frac{1}{2}\mathbf{e}^T{}_{c,i}\bar{\mathbf{M}}_i\mathbf{e}_{c,i} . \tag{25}$$

Differentiating (25) and substituting (24) yields,

$$\dot{V}_{2,i}(\mathbf{e}_{c,i}) = \mathbf{e}^T{}_{c,i}[\bar{\tau}_i - (\bar{B}_i\mathbf{v}_i + \bar{\mathbf{M}}_i\dot{\mathbf{v}}_{c,i} + \bar{\tau}_{d,i})] + \frac{1}{2}\mathbf{e}^T{}_{c,i}\dot{\bar{\mathbf{M}}}_i\mathbf{e}_{c,i} . \tag{26}$$

Applying all the Properties and Assumption, we obtain,

$$\begin{aligned}\dot{V}_{2,i}(\mathbf{e}_{c,i}) &\le \mathbf{e}^T{}_{c,i}\bar{\tau}_i + \|\mathbf{e}_{c,i}\|\{\|\bar{B}_i\|\|\mathbf{v}_i\| + \|\bar{\mathbf{M}}_i\|\|A_{1,i}\|\|\mathbf{v}_{c,i}\| \\ &+ \|\bar{\mathbf{M}}_i\|\|A_{2,i}\|\|\mathbf{v}_{r,i}\| + \|\bar{\mathbf{M}}_i\|\|A_{3,i}\|\|\dot{\mathbf{v}}_{r,i}\| + \|\tau_{d,i}\| + \|\mathbf{F}_K(q_i,q_j)\|\} \\ &= \mathbf{e}^T{}_{c,i}\bar{\tau}_i + \|\mathbf{e}_{c,i}\|\delta_i^T\xi_i\end{aligned} \tag{27}$$

where,

$$\begin{aligned}\delta_i^T &= \{\|\bar{\mathbf{M}}_i\|\|A_{1,i}\|, \|\bar{B}_i\|, \|\bar{\mathbf{M}}_i\|\|A_{2,i}\|, \|\bar{\mathbf{M}}_i\|\|A_{3,i}\|, \|\tau_{d,i}\|, 1\} \\ \xi_i^T &= \{\|\mathbf{v}_{c,i}\|, \|\mathbf{v}_i\|, \|\mathbf{v}_{r,i}\|, \|\dot{\mathbf{v}}_{r,i}\|, 1, \|\mathbf{F}_{K,i}(q_i,q_j)\|\}\end{aligned} \tag{28}$$

Here $\delta_i$ is bounded by the above properties and assumptions, and $\xi_i$ is a known, positive definite vector. Hence, in order to make (27) negative definite, choose,

$$\bar{\tau}_i = -K_i\mathbf{e}_{c,i}\|\xi_i\|^2 , \tag{29}$$

where $K_i = \begin{bmatrix} K_{1,i} & 0 \\ 0 & K_{2,i} \end{bmatrix}$ is the matrix control gain and $K_{1,i}$, $K_{2,i}$ are positive constants. The control input is then,

$$\tau_i = -(\mathbf{S}_i{}^T\mathbf{E}_i)^{-1}K_i\mathbf{e}_{c,i}\|\xi_i\|^2 . \tag{30}$$

Using the Lyapunov function $V_i = V_{1,i} + V_{2,i}$ [11, 16], $\mathbb{C}_i = \begin{bmatrix}\mathbf{e}_i & \mathbf{e}_{c,i}\end{bmatrix}^T$ is globally uniformly bounded and the velocity tracking error becomes arbitrarily small by increasing the control gain $K_i$.

Note that the reference velocity vector is included in the control input since the motion controller could provide time varying reference velocities. The compliant frame force $\mathbf{F}_{K,i}(q_i,q_j)$ is also taken into consideration by the controller. As the configuration and environment become more complicated, however, the controller without the compliant frame force compensation may be desirable (i.e., $\mathbf{F}_{K,i}(q_i,q_j)$ could be considered as a disturbance) [14].

*D. Multi-axle distributed control design*

The distributed controller is designed for a multi-axle CFMMR based on the above single-axle controller, i.e., the distributed controller is composed of $n$ independent controllers $\tau_j, j = 1,...,n$ as:

$$\tau_j = -(\mathbf{S}_j{}^T\mathbf{E}_j)^{-1}K_j\mathbf{e}_{c,j}\|\xi_j\|^2 . \tag{31}$$

*Proposition: The multi-axle CFMMR can achieve stable trajectory tracking with the distributed controller (31) if the response of each module is globally uniformly bounded by its corresponding single-axle controller.*

*Proof*: Choose the composite Lyapunov function candidate,

$$\begin{aligned}V &= V_1 + ... + V_i + ... + V_n \\ &= V_{1,1} + V_{2,1} + ... + V_{1,i} + V_{2,i} + ... + V_{1,n} + V_{2,n}\end{aligned} . \tag{32}$$

Differentiating (32) and applying (19) and (29) yields,

$$\begin{aligned}\dot{V} &\le -k_{X,1}e^2{}_{X,1} - k_{Y,1}\sin^2 e_{\phi,1} - K_{1,\max}\left\{\|\mathbf{e}_{c,1}\|\|\xi_1\| - \frac{\|\delta_1\|}{2K_{1,\max}}\right\}^2 \\ &+ \frac{\|\delta_1\|^2}{4K_{1,\max}} - ... - k_{X,i}e^2{}_{X,i} - k_{Y,i}\sin^2 e_{\phi,i} \\ &- K_{i,\max}\left\{\|\mathbf{e}_{c,i}\|\|\xi_i\| - \frac{\|\delta_i\|}{2K_{i,\max}}\right\}^2 + \frac{\|\delta_i\|^2}{4K_{i,\max}} - ... - k_{X,n}e^2{}_{X,n} \\ &- k_{Y,n}\sin^2 e_{\phi,n} - K_{n,\max}\left\{\|\mathbf{e}_{c,n}\|\|\xi_n\| - \frac{\|\delta_n\|}{2K_{n,\max}}\right\}^2 + \frac{\|\delta_n\|^2}{4K_{n,\max}}\end{aligned}$$

where $K_{1,\max},...K_{n,\max}$ are maximum elements of positive definite matrices $K_1,...K_n$ individually, $k_{X,1},...k_{X,n}$ are positive constants, and $\|\delta_1\|,...\|\delta_n\|$ are bounded. Therefore,

$$\begin{aligned}\dot{V} &\le -k_{X,1}e^2{}_{X,1} - k_{Y,1}\sin^2 e_{\phi,1} - ... - k_{X,i}e^2{}_{X,i} - k_{Y,i}\sin^2 e_{\phi,i} \\ &- ... - k_{X,n}e^2{}_{X,n} - k_{Y,n}\sin^2 e_{\phi,n} \\ &= -W(\mathbf{e})\end{aligned} \tag{33}$$

when,

$$\|\mathbf{e}_{c,1}\| \ge \frac{1}{K_{1,\max}}\frac{\|\delta_1\|}{\|\xi_1\|}, ..., \|\mathbf{e}_{c,n}\| \ge \frac{1}{K_{n,\max}}\frac{\|\delta_n\|}{\|\xi_n\|} , \tag{34}$$

where $\mathbf{e} = \begin{bmatrix}\mathbb{C}_1 & ... & \mathbb{C}_n\end{bmatrix}^T$, $\mathbb{C}_j = \begin{bmatrix}\mathbf{e}_j & \mathbf{e}_{c,j}\end{bmatrix}^T, j = 1,...,n$, and $W(\mathbf{e})$ is a continuous positive definite function. Hence we conclude $\mathbf{e}$ is globally uniformly bounded [16] and can be minimized by tuning controller gains. Δ

## V. Controller Interaction

*A. Cascade connection*

The cascade connection is developed to generate the reference velocities $\mathbf{v}_r$ of each axle based on the simplified kinematics to feed into the robust *dynamic motion controller*, Fig. 4. The linear and angular velocities of each axle can be described from the linear velocity $v$ and path curvature $\kappa$ of the center posture, $O$. By the assumption of pure bending, where $\psi = \psi_1 = -\psi_2$, $\psi$ may be solved numerically using the expression for the path curvature of point $O$, Fig. 2:

$$\kappa = \frac{1}{r} = \frac{2\psi}{L\cos\psi} , \quad (35)$$

where $L$ is the frame length. Referring to the foreshortening result of the frame presented in [2, 12], the linear and angular velocities of each axle, $v_i$ and $\omega_i$ are,

$$\begin{aligned} v_i &= \frac{v}{\cos\psi} + \frac{(-1)^i}{6} L\psi\dot{\psi} \\ \omega_i &= \dot{\phi} + (-1)^{i-1}\dot{\psi} . \end{aligned} \quad \begin{cases} i = 1 \text{ for front axle} \\ i = 2 \text{ for rear axle} \end{cases} \quad (36)$$

where $d$ is the half length of an axle and $r_w$ is the wheel radius.

### B. Feedback

We consider three different controller interaction structures here; (1) the *kinematic motion controller* is driven by ideal unicycle kinematics and the actual robot positions and velocities are only fed back to the *dynamic motion controller*; (2) the actual robot velocities are fed back to both the *kinematic motion controller* and the *dynamic motion controller*, and the actual robot positions are fed back to only the *dynamic motion controller*; (3) the actual robot positions are fed back to both the *kinematic motion controller* and the *dynamic motion controller*, and the actual robot velocities are only fed back to the *dynamic motion controller*, Fig. 4. We will evaluate each form of controller interaction in the next section.

## VI. Controller Evaluation

### A. Methods and Procedures

Experiments for the distributed robust control evaluation were conducted on a two-module CFMMR experimental platform, Fig. 1, at the University of Utah operating on a smooth, flat, high traction carpet surface. Experimental results presented here examine the controller interaction structures discussed in Sect. V to evaluate their effect on posture regulation performance.

The robot is controlled via tether by a dSpace™ 1103 DSP board and an external power supply. Each wheel is actuated by a DC motor where the real-time position of each wheel is detected by an encoder and odometry is used for predicting axle posture. Second order filters were initially used to filter encoder signal noise, but this caused marginal stability due to phase lag. Hence encoder signals are directly used to provide position and velocity feedback.

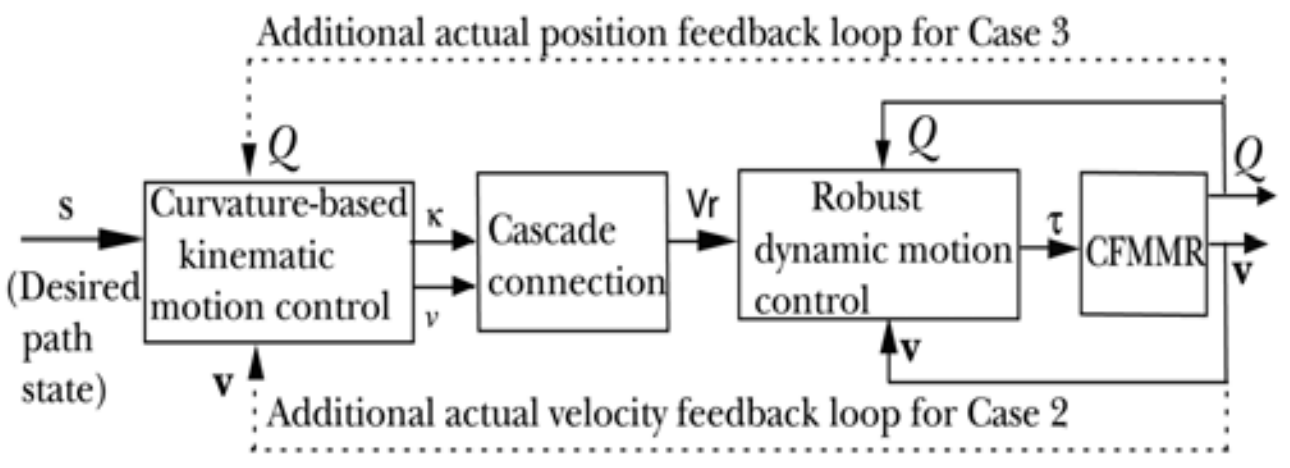

Fig. 4. Controller interaction structures

### B. Results &Discussion

Fig. 5 (a)-(c) shows the angular velocities of the rear axle for Cases 1-3. As Fig. 5 indicates, Case 2 filtered sensor noise the best, and Case 3 has oscillations when the robot was close to the origin at the final time. Fig. 6 (a)-(c) shows experimental posture regulation for Cases 1-3, where the trajectory of the midpoint, $O$, of the robot is represented by the solid line, the trajectory of the wheels are represented by dashed lines and the target final position of the midpoint $O$ is indicated by dotted coordinate lines. As Fig. 6 indicates, though, Case 2 caused the largest posture regulation error and Cases 1 and 3 have small posture regulation errors.

As Fig. 5 and Fig. 6 indicate, the system performed well for posture regulation according to odometry, even with large sensor noise. Compared with the pure model-based back-stepping controller presented in [2], the controller derived here decreases off-tracking without tuning control gains [14].

Table 1 shows the final position error of the midpoint, $O$, according to odometry and actual measurements. Variables $e$, $\theta$ and $\alpha$ are from the *kinematic motion controller*; *Xo-d* and *Yo-d* are the errors from the *dynamic motion controller* relative to the reference provided by the kinematic controller; $X_O$ and $Y_O$ are the final position error according to odometry; $X_A$ and $Y_A$ are the measured position error of the robot, and DEV represents the radial distance between the measured and odometric data.

Case 1 has the smallest $e$, $\theta$ and $\alpha$ since it is based on an ideal kinematic model. Small *Xo-d* and *Yo-d* show that the dynamic controller follows the reference from the kinematic controller quite well. Thus, Case 1 has small final position error according to both odometry and actual measurements.

The additional velocity feedback (Case 2) between the kinematic motion control and dynamic motion control laws causes the control system to have much better noise rejection, but largest finite error since sensor noise is added to the input of the *kinematic motion controller* and deviations from the reduced kinematic model are inevitable. This disturbs the kinematic controller and increases both kinematic controller error (i.e., $e$, $\theta$ and $\alpha$) and dynamic controller error (i.e., *Xo-d* and *Yo-d*). As a result, the net steady state posture error becomes larger than Case 1, but deviations (i.e. DEV) from actual error measurements are comparable.

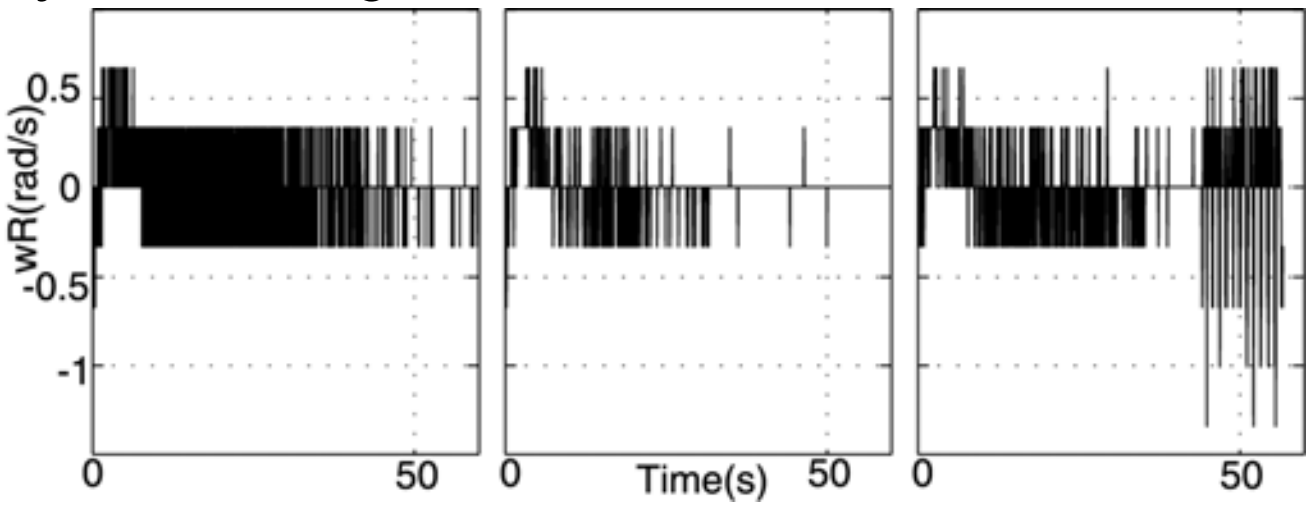

(a) Case 1 (b) Case 2 (c) Case3

Fig. 5. Angular velocities of rear axle for posture regulation.

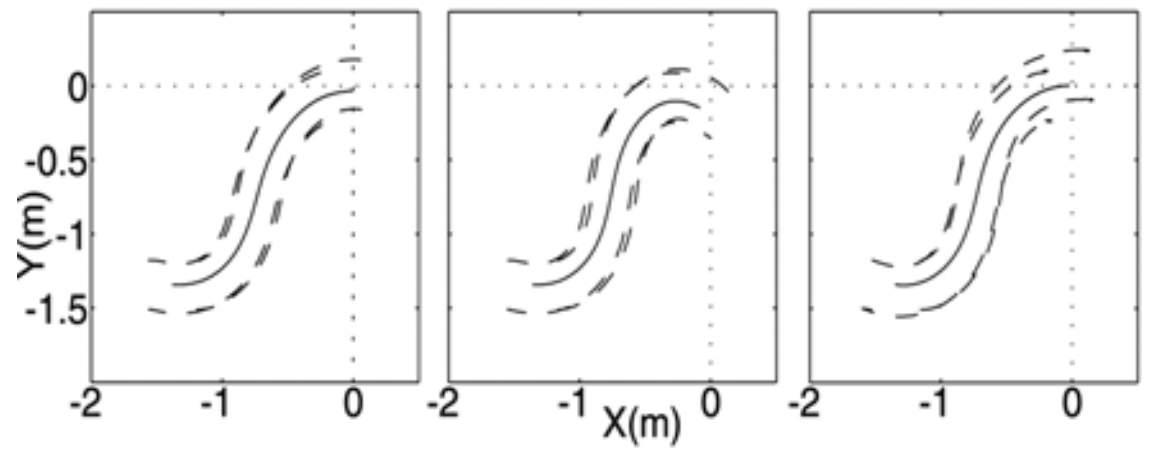

(a) Case 1 (b) Case 2 (c) Case 3

Fig. 6. Experimental posture regulation according to wheel odometry.

According to odometry, the additional position feedback of Case 3 achieves the smallest position error. Actual measurements indicate that the position error is slightly larger than that of Case 1. The most important point to note, however, is that the DEV for Case 3 is the largest, which is an indicator that Case 3 is the most prone to wheel slip. This may be the result of gentle oscillation near the origin due to the interaction between the kinematic controller and dynamic controller.

Since odometry is subject to measurement errors, additional sensor algorithms must be developed to compensate drift. At this point, Case 1 is most pertinent for posture regulation accuracy, while Case 2 shows improved noise rejection and reduced wheel slip.

## VII. CONCLUSIONS

This paper introduces a novel control algorithm combining a curvature-based kinematic motion controller and a robust dynamic controller for Wheeled Compliant Framed Modular Mobile Robots. Experimental results for a two-axle CFMMR configuration demonstrate the efficiency and robustness of the proposed control technique. This control algorithm is generally applicable to other cooperative mobile robots, which have unknown or partially known uncertainties. Future work will focus on additional sensor algorithms, controller and sensor algorithm interaction, and the behavior of more than three CFMMR modules.

Table 1. Experimental position errors.

| | *According to wheel odometry* | | | | | | | | *Actual measurement* | | | |
|---|---|---|---|---|---|---|---|---|---|---|---|---|
| Case | *e(m)* | $\theta$ *(rad)* | $\alpha$*(rad)* | *Xo-d(m)* | *Yo-d(m)* | *Xo(m)* | *Yo (m)* | $\lVert Xo, Yo \rVert$ | $X_A$ *(m)* | $Y_A$ *(m)* | $\lVert X_A, Y_A \rVert$ | *DEV* |
| Case1 | 0.004 | 0.005 | -0.004 | 0.0025 | -0.030 | -0.003 | -0.030 | 0.030 | 0.089 | -0.055 | 0.100 | 0.095 |
| Case2 | 0.100 | -0.15 | 0.46 | 0.013 | 0.019 | -0.085 | -0.145 | 0.168 | -0.005 | -0.180 | 0.180 | 0.087 |
| Case3 | -0.020 | -0.01 | 0.25 | -0.005 | 0.0005 | -0.026 | 0.0015 | 0.026 | 0.100 | -0.060 | 0.110 | 0.140 |